\pdfoutput=1

\documentclass[11pt]{article}
\usepackage{authblk}
\usepackage[]{acl}

\usepackage{times}
\usepackage{latexsym}

\usepackage[T1]{fontenc}

\usepackage[utf8]{inputenc}

\usepackage{microtype}

\usepackage{times}
\usepackage{latexsym}
\usepackage{booktabs}

\usepackage{amsmath}
\usepackage{url}
\usepackage{mathdots}
\usepackage{yhmath}
\usepackage{cancel}
\usepackage{color}
\usepackage{siunitx}
\usepackage{array}
\usepackage{multirow}
\usepackage{amssymb}
\usepackage{gensymb}
\usepackage{graphicx}
\usepackage{float}
\usepackage{rotating}
\usepackage{tabularx}
\usepackage{booktabs}
\usepackage{svg}
\usepackage{tablefootnote}
\usepackage{cleveref}
\usepackage{tabularx}
\usepackage{xcolor}
\usepackage{makecell}
\usepackage{enumitem}
\usepackage{color, colortbl}
\usepackage{subcaption}

\definecolor{Gray}{gray}{0.9}

\renewcommand{\sectionautorefname}{\S\kern-0.2em}
\renewcommand{\subsectionautorefname}{\S\kern-0.2em}
\renewcommand{\subsubsectionautorefname}{\S\kern-0.2em}

\newcommand{\newpara}[1]{\vspace{0.3em} \noindent \textbf{#1}~ \hspace{0.4em}}

\newcommand{\ours}{\textsc{POKI}}

\definecolor{mygreen}{HTML}{dfead5}
\definecolor{myred}{HTML}{ffe7e6}

\makeatletter
\renewcommand*{\@fnsymbol}[1]{\ensuremath{\ifcase#1\or \dagger\or \ddagger\or
    \mathsection\or \mathparagraph\or \|\or **\or \dagger\dagger
    \or \ddagger\ddagger \else\@ctrerr\fi}}
\makeatother

\title{Achieving Conversational Goals with\\Unsupervised Post-hoc Knowledge Injection
}

\author[$\clubsuit$]{\textbf{Bodhisattwa Prasad Majumder}}
\author[$\diamondsuit$]{\textbf{Harsh Jhamtani}}
\author[$\clubsuit$]{\textbf{\qquad \qquad \qquad \qquad \qquad Taylor Berg-Kirkpatrick}}
\author[$\clubsuit$]{\textbf{Julian McAuley}}
\affil[$\clubsuit$]{Department of Computer Science and Engineering, UC San Diego \protect\\ \tt \{bmajumde, tberg, jmcauley\}@eng.ucsd.edu}
\affil[$\diamondsuit$]{School of Computer Science, Carnegie Mellon University \protect\\ \tt jharsh@cs.cmu.edu}

\makeatletter
\renewcommand\outauthor{
    \begin{tabular}[t]{>{\centering}p{14cm}} 
    \bf\@author
    \end{tabular}}
\makeatother

\begin{document}
\maketitle

\begin{abstract}
A limitation of current neural dialog models is that they tend to suffer from a lack of specificity and informativeness in generated responses, primarily due to dependence on training data that covers a limited variety of scenarios and conveys limited knowledge.
One way to alleviate this issue is to extract relevant knowledge from external sources at decoding time and incorporate it into the dialog response.
In this paper, we propose a post-hoc knowledge-injection technique where we first retrieve a diverse set of relevant knowledge snippets conditioned on both the dialog history and an initial response from an existing dialog model. We construct multiple candidate responses, individually injecting each retrieved snippet into the initial response using a gradient-based decoding method, and then select the final response with an unsupervised ranking step.
Our experiments in goal-oriented and knowledge-grounded dialog settings demonstrate that human annotators judge the outputs from the proposed method to be more engaging and informative compared to responses from prior dialog systems. We further show that knowledge-augmentation promotes 
success in achieving conversational goals in both experimental settings.
\end{abstract}

\section{Introduction}

\begin{figure}[t!]
    \centering
    \includegraphics[trim=240 325 645 255,clip, width=0.95\linewidth]{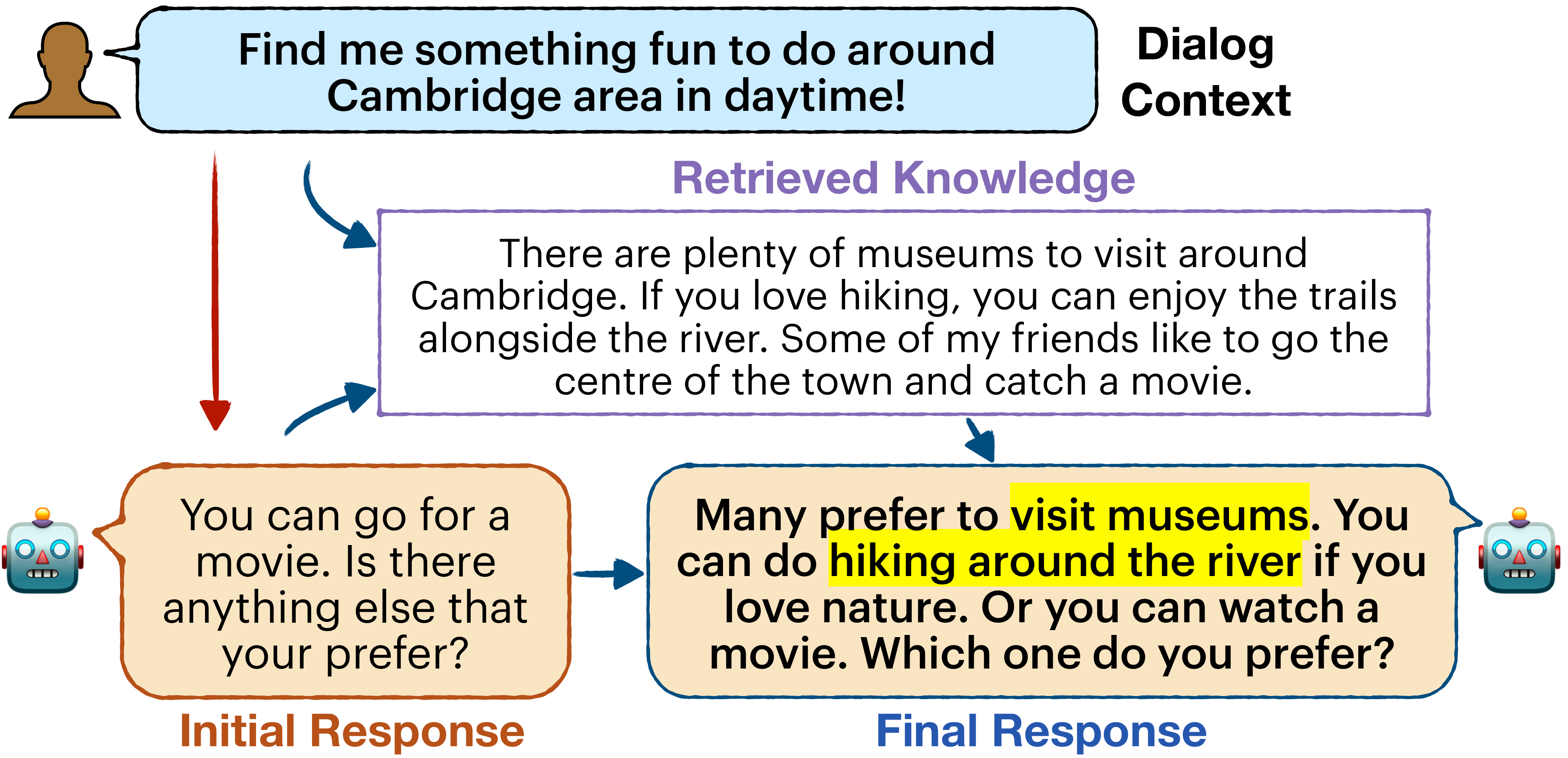}
    \caption{\small
    Augmenting initial response from an existing dialog model with relevant external knowledge leads to more \textit{engaging} and \textit{informative} responses improving the success in achieving the conversational goal (here, finding a fun activity). 
    }
    \label{fig:overview}
    \vspace{-0em}
\end{figure}

Generic responses which lack specificity have been a major issue in existing dialog models \cite{DBLP:conf/nips/Hosseini-AslMWY20, DBLP:conf/iclr/DinanRSFAW19}. 
The issue 
in part stems from bottlenecks in dialog models due to a limited scope of scenarios and access to limited knowledge available during training.
On the other hand, encoding all possible world knowledge at training time is not feasible, and even undesirable in cases where knowledge sources are dynamically varying \cite{DBLP:conf/aaai/GhazvininejadBC18, DBLP:conf/emnlp/MajumderLNM20, DBLP:conf/emnlp/ZhaoWXTZY20, DBLP:conf/kdd/BruynLBD20, DBLP:conf/iclr/KimAK20, DBLP:conf/naacl/PrabhumoyeHZBS21}.
One possible approach is to incorporate relevant knowledge at decoding-time.
For example, in \autoref{fig:overview}, the user is seeking options for a fun activity around Cambridge. While the initial dialog response suggests watching a movie as an option, it does not provide any information behind that choice. 

\begin{figure*}[t!]
    \centering
    \includegraphics[trim=10 360 10 270,clip, width=\linewidth]{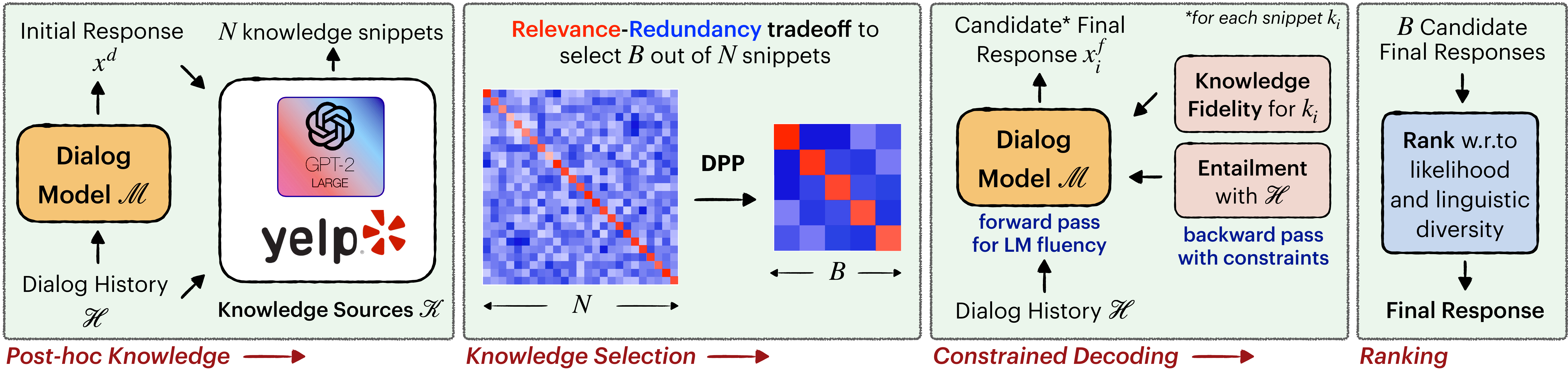}
    \vspace{-1.2em}
    \caption{\small 
    Pipeline of \ours{}: It first retrieves \textbf{post-hoc knowledge} from external sources based on dialog history and an initial response from a dialog model. Then the most relevant and diverse knowledge snippets are \textbf{selected} from the retrieved set. Each selected snippet is individually combined with the initial response through \textbf{constrained decoding} to generate a candidate final response. At last, the final response is selected via an unsupervised \textbf{ranking} step. Note that \ours{}~requires no additional training.
    }
    \label{fig:model}
    \vspace{0em}
\end{figure*}



We propose and evaluate an approach for unsupervised knowledge injection into a dialog model's response at decoding time\footnote{Code:~\url{https://github.com/majumderb/poki}}---not addressed in any previous work. We first sample a response from the model (trained on dialog data) conditioned on the dialog context. Next, we utilize the dialog context and the sampled response to query external knowledge sources. Finally, the retrieved knowledge is used to construct a more informative and engaging response (\Cref{fig:overview}).
A major advantage of such post-hoc knowledge injection is its flexibility in adding newer knowledge sources especially where the success of achieving conversational goals relies upon the availability of relevant knowledge.
Post-hoc injection also promotes efficiency in NLP applications \cite{DBLP:journals/cacm/SchwartzDSE20,DBLP:conf/acl/StrubellGM19}: it mitigates the need to retrain dialog models to accommodate dynamically evolving knowledge. 

We experiment with two types of knowledge sources: language models, which we treat as parametric knowledge bases \cite{DBLP:conf/emnlp/PetroniRRLBWM19, DBLP:conf/nips/BrownMRSKDNSSAA20}; and user review datasets such as Yelp reviews \cite{DBLP:journals/corr/HajasGK14} as non-parametric knowledge sources (\autoref{sec:post-hoc-knowledge}). Since it is possible to gather a large amount of related knowledge given a query, we select a relevant and diverse (estimated via information-theoretic measures) subset of knowledge snippets using an unsupervised method (\autoref{sec:know-select}). 
Then, a gradient-based inference approach is used to construct an updated response that incorporates the selected knowledge (\autoref{sec:decoding}).
Note that our framework does not require retraining the existing dialog model---it only relies upon updating the model's output hidden states at decoding time for unsupervised knowledge injection.


We experiment with two scenarios: goal-oriented and knowledge-grounded dialog where the training data covers only a fraction of the needed knowledge. Automatic evaluation reveals that our method is capable of generating highly diverse responses in both settings. In some cases, the generated response shows high overlap with the original target response showing that our unsupervised method bridges the knowledge gap between available knowledge and human-written responses present in the existing dialog corpus. An extensive human evaluation confirms that generated responses are indeed engaging, interesting, and human-like without any loss in fluency.

To pinpoint the usefulness of knowledge injection in the above settings, we design a real-time study (\autoref{sec:user-study}) where users interact with our system to reach a conversational goal (e.g.~planning a holiday or knowing more about the solar system). We find that external knowledge enables users to achieve their goals more efficiently.
Additionally, we observe that the our approach of sub-selecting relevant but diverse knowledge leads to responses that promote success in achieving conversational goals.
\section{Post-hoc Knowledge for Dialog}
\label{sec:post-hoc-knowledge}
Our goal is to construct a dialog response by injecting knowledge (from external textual sources) at decoding time, without having to retrain the models. 
Consider a dialog model $\mathcal{M}$ from which we can sample a dialog response $x^d$ given a dialog history $\mathcal{H}$.
We shall refer to the response $x^d$ sampled from such a model without any decoding time knowledge injection as the \emph{initial} response.

However, as motivated earlier, samples from such a dialog model 
often 
lack
detail. To improve 
such responses, we retrieve and incorporate relevant external knowledge $k$ into the initial response.
To achieve our goal, we construct a \textit{query} using both dialog history $\mathcal{H}$ and the initial response $x^d$, and gather a relevant knowledge candidate $k$ from a knowledge source $\mathcal{K}$. The retrieved snippet can 
provide useful information to the end-user to achieve the conversational goal (see \autoref{sec:user-study}). We explore both parametric (e.g~querying a language model) and non-parametric (e.g.~deterministic retrieval using word-overlap) ways to  obtain post-hoc knowledge. 


\subsection{Parametric knowledge sources}
Pretrained language models (PTLM) are typically trained with a vast amount of text that spans a diverse range of domains. \citet{DBLP:conf/emnlp/PetroniRRLBWM19, DBLP:conf/nips/BrownMRSKDNSSAA20} showed that such PTLMs can be used as a source of knowledge when queried with suitable textual prompts (e.g.~\textit{Seattle is famous for} \rule{0.5cm}{0.15mm}). 
To use PTLMs in our use-case, we construct useful prompts from dialog history and the \textit{initial} response. We assemble simple prompts inspired from various knowledge-seeking situations in dialog \cite{DBLP:conf/emnlp/ShwartzWBBC20} such as \texttt{[KP]} \textit{is famous for} \rule{0.5cm}{0.15mm}, \textit{Here is what I know about} \texttt{[KP]}: \rule{0.5cm}{0.15mm}, where \texttt{[KP]} is a key-phrase\footnote{It possible that a lack of key-phrases results in no knowledge. Key-phrase extraction details are in \autoref{ap:code}.
} extracted from dialog context. We use \texttt{gpt2-large} as the PTLM. For example, a query ``Here is what I know about fun things around Cambridge:" results in ``\textit{There are
plenty of museums to visit around Cambridge. If you love hiking, you can enjoy the trails alongside the river...}" as shown in \autoref{fig:overview}. A complete list of prompts is provided in
\autoref{ap:code}. We finally rank each knowledge snippet $k$ using the likelihood obtained from the PTLM for a concatenated input of $k$ and dialog history and choose 
the
most likely.

\subsection{Non-parametric knowledge sources}
External knowledge in the form of a text corpus can be used as a non-parametric knowledge source available at decoding time. Compared to parametric knowledge sources, such sources do not generate text as knowledge snippets, but 
offer the advantage of 
high quality and reliability of human written text.
We consider the dialog history
and the \emph{initial} response as 
a 
query to retrieve relevant knowledge instances from the corpus. Next, we identify the top relevant instances in the given corpus with respect to the constructed query using cosine similarity on TF-IDF based representations \cite{DBLP:journals/ipm/RobertsonWH95}.

\section{Unsupervised Knowledge Injection in Generated Dialog}

\label{sec:gradinfer}

Effectively utilizing the retrieved knowledge snippets to construct an enriched dialog response encompasses two major challenges. Firstly, it is not practical to use potentially hundreds of knowledge snippets obtained from the retrieval step for
a single response generation. Thus, we need to find a relevant but diverse subset of the snippets. Secondly, the dialog model $\mathcal{M}$ is trained to condition only on the dialog context, and not on the external knowledge. Hence, to leverage the knowledge snippets, we need a decoding strategy to rewrite the initial response $x^d$ such that the resulting final response $x^f$ should closely follow the knowledge snippet to be injected without a loss in the fluency and consistency. Thus, our method requires no additional training and only assumes a language model trained on dialog context (i.e.~$\mathcal{M}$). We refer to our proposed framework (\autoref{fig:model}) as \textbf{\ours}~(\textbf{Po}st-hoc \textbf{K}nowledge \textbf{I}njection in Generated Dialog).

\subsection{Relevance-Redundancy Tradeoff for Knowledge Selection}
\label{sec:know-select}

At each turn, we obtain $N$ knowledge snippets from both the parametric and non-parametric sources. We wish to select a subset of $B$ (out of $N$) relevant but diverse knowledge snippets. 

We define relevance score of a snippet $k_i$ with respect to the dialog history $H$ using pointwise mutual information (PMI) as follows: 
\[
\mathbb{REL}_i = \operatorname{PMI}(k_i, \mathcal{H}) = \log\left(\frac{p(\mathcal{H}|k_i)}{p(\mathcal{H})}\right),
\] 
Thus, a high PMI score would imply a larger semantic similarity between the snippet $k_i$ and $H$.  
%
To account for redundancy
between the snippet pair $k_i$, $k_j$ 
we again use
the PMI score as follows: 
\[
\mathbb{RED}_{ij, j>i} = \operatorname{PMI}(k_i, k_j) = \log\left(\frac{p(k_j|k_i)}{p(k_j)}\right).
\]
The redundancy score is symmetric i.e. $\mathbb{RED}_{ij} = \mathbb{RED}_{ji}$ as $\operatorname{PMI}$ is a symmetric measure.

We estimate probabilities (both conditional and marginal) $p(.)$ in the above equations using GPT2 language model, following past work \cite{DBLP:conf/eacl/PadmakumarH21}. The PMI measure is often considered better than other n-gram-based overlap metrics to measure the degree of association between two sentences \cite{DBLP:conf/emnlp/KedzieMD18, DBLP:conf/eacl/PadmakumarH21}. Semantically similar 
phrases occur in both sentences that 
can easily be ignored by
overlap based metrics.

\paragraph{Selection via Determinantal Point Processes.} To select $B$ knowledge snippets out of $N$ with a relevance-redundancy trade-off, we use a subset selection process named Determinantal Point Process (DPP) \cite{DBLP:conf/icml/KuleszaT11}. DPP employs a non-uniform selection
that assigns low probability to subsets (here, of knowledge snippets) that are less diverse by modeling the repulsive correlation between independently occurring datapoints (see \autoref{fig:model}).

We build an $N \times N$ kernel matrix $\mathcal{D}$, which is real, symmetric and positive semi-definite. The diagonal entries $\mathcal{D}_{ii}$ are populated by the squared relevance score of the $i$-th knowledge $\mathbb{REL}_i$ and the off-diagonal entries $\mathcal{D}_{ij}$ are $\beta \,\times$ squared redundancy scores $\mathbb{RED}_{ij}$. We adjust $\beta$ in such a way that $\mathcal{D}$ always remains positive semi-definite (more details in \cite{DBLP:conf/cikm/WilhelmRBJCG18}). To select a subset of $B$, a DPP assigns a probability of sampling such a subset proportional to the determinant of the submatrix $\mathcal{D}_B$ of $\mathcal{D}$, constructed using the indices of the subsetted items. The DPP probability is geometrically related to the
volume of the parallelepiped spanned by the selected knowledge snippets. Diverse knowledge snippets tend to be orthogonal in their space hence span larger volume \cite{DBLP:journals/ftml/KuleszaT12}.

Choosing $B$-size submatrix from $N$-size $\mathcal{D}$ is a combinatorial problem and can become prohibitively costly when $N$ is very high. Hence, we use a greedy method \cite{DBLP:conf/cikm/WilhelmRBJCG18} where we initialize the selection with the most relevant $k_i$ and subsequently select the next $k_j$
that maximizes the determinant of the resultant submatrix.

\subsection{Gradient-based Constrained Decoding for Knowledge Injection}
\label{sec:decoding}
Upon selecting $B$ knowledge snippets, we want to individually inject each knowledge snippet into $x^d$ to construct a candidate final response $x^f$ at inference time.

Previous works have addressed the problem of unsupervised modification of already-generated text using gradient-based decoding \cite{DBLP:conf/iclr/DathathriMLHFMY20,DBLP:conf/emnlp/QinSWBHBBC20} that 
employs an iterative procedure consisting of a forward 
and a backward pass. The forward pass on the generative model
(here, $\mathcal{M}$)
encourages fluency of the generated text while the backward pass performs
gradient ascent on certain desired constraints.
%
Note that due to the discrete nature of $x_d$, it is not possible to directly update it via back-propagation. Therefore, we maintain the sequence of hidden representations of 
each output token as $z$ from the dialog model. Each output token $x^d_{(t)}$ is realized via~$p(x^d_{(t)}) \sim \operatorname{softmax}(Wz_{(t)}/\tau)$, where $\tau$ is the temperature hyperparameter, $W$ is the output embedding matrix (shared with the input), and $Wz_{(t)} \in \mathcal{R}^V$ ($V$ is the size of the vocabulary).

\newpara{Constraints.}
Following \citet{DBLP:conf/acl/MajumderBMJ20}, we define a \textbf{knowledge fidelity} objective that encourages 
$x^f$ to be minimally different from the knowledge snippet $k$. 
We achieve this by minimizing the cross entropy loss ($\operatorname{CE}$) between knowledge tokens $k_{(1)},\ldots,k_{(T)}$ as labels and $Wz_{(1)},\ldots,Wz_{(T)}$ as the logits.

We further notice that injected knowledge can influence the generation in such a way that it contradicts with responses uttered during previous turns. Hence, we also want 
$x^f$ to be entailed with the dialog history $\mathcal{H}$. 
We build an \textbf{entailment} classifier $\theta(z,\mathcal{H})$ that predicts the probability of $x^f$ (ideally, the hidden representation $z$ of $x^f$) entailing $\mathcal{H}$. 
The classifier $\theta(z,\mathcal{H})$ is a bag-of-words classification layer with hidden states $z$ from $\mathcal{M}$ and fine-tuned using the DNLI dataset \cite{DBLP:conf/acl/WelleckWSC19} to predict whether the current response is entailed with previous responses or not.

\newpara{Decoding.} In the subsequent forward and backward passes, the hidden representation $z$ is gradually perturbed via gradient ascent on the respective objectives. During backward pass, the objective with constraints is
\[
\mathcal{L}(\mathcal{H},k;z) = \alpha \log \theta(z,\mathcal{H})
    - \lambda  \operatorname{CE}(k,Wz)
\]    
with hyperparameters $\alpha$ and $\lambda$.
We use back-propagation to update $z$ with the gradient $\nabla_{z} \mathcal{L}(\mathcal{H},k;z)$ while the parameters of $\mathcal{M}$ remain fixed. The updated latent representations of $z$ after the backward pass are denoted as $z^\mathit{bw}$.

A forward pass with $\mathcal{M}$ is required to 
regularize the hidden states $z$ toward the original dialog model objective to obtain $z^\mathit{fw}$.
Corresponding to the $t^{\text{th}}$ token, the hidden states for 
the
${t+1}^{\text{th}}$ time step are computed 
via a weighted addition of backward and forward hidden states, i.e., $z_{(t+1)} = \gamma \times z^\mathit{bw}_{(t)} + (1-\gamma) \times z^\mathit{fw}_{(t)}$ where $\gamma\in(0,1)$ is a hyperparameter. 

During generation, we start by sampling the initial response $x^d$ with greedy decoding
from $\mathcal{M}$. 
The hidden states $z$ (of $x^d$) are iteratively updated by alternate
backward and forward passes. 
The final response is sampled as $x^f \sim \operatorname{softmax}(Wz/\tau)$. The number of iterations ($=5$) and the $\gamma$ ($=0.45$) were chosen by maximizing the Z-normalized sum of dialog model perplexity and linguistic diversity (\% of distinct bigrams) in a greedy hyperparameter search.
More details are in \autoref{ap:code}.

\begin{table*}[t!]
\centering
\footnotesize
\begin{minipage}{0.48\textwidth}
\centering
\begin{tabular}
{@{}lccccc@{}}
\toprule
\bf System &  \bf \hspace{-3mm}Acc \bf & \bf \hspace{-2mm}BLEU & \bf \hspace{-2mm}BRTSc & \bf \hspace{-2mm}D-2 & \bf \hspace{-2mm}ENTR\\
\midrule
KCopy & \hspace{-3mm}70.1 & \hspace{-3mm}4.1 & 
\hspace{-3mm}62.3 & \hspace{-3mm}3.16 &  \hspace{-2mm}2.41 \\
SimpleTOD \shortcite{DBLP:conf/nips/Hosseini-AslMWY20} & \hspace{-3mm}70.1 & \bf \hspace{-3mm}15.0 & 
\bf \hspace{-3mm}79.2 & \hspace{-2mm}0.56 & \hspace{-2mm}0.90\\
SimpleTOD+ \shortcite{DBLP:journals/corr/abs-2010-12757} & \hspace{-3mm}69.8 & \hspace{-3mm}12.1 &
\hspace{-3mm}68.1 & \hspace{-2mm}0.81 & \hspace{-2mm}1.11\\
Arranger \shortcite{DBLP:journals/corr/abs-2010-12757} & \hspace{-3mm}70.2 & \hspace{-3mm}12.3 &
\hspace{-3mm}68.5 & \hspace{-2mm}0.93 & \hspace{-2mm}1.15\\
Rewriter \shortcite{DBLP:journals/corr/abs-2010-12757} & \hspace{-3mm}70.2 & \hspace{-3mm}12.1 &
\hspace{-3mm}69.4 & \hspace{-2mm}1.03 & \hspace{-2mm}1.45\\
\ours & \bf \hspace{-3mm}71.1 & \hspace{-3mm}13.7 &
\hspace{-3mm}74.5 & \bf \hspace{-2mm}3.78 & \bf \hspace{-2mm}2.67\\
\hspace{0.6em} w/o Entailment & \hspace{-3mm}69.9 & \hspace{-3mm}10.9 &
\hspace{-3mm}67.8 & \bf \hspace{-2mm}3.67 & \bf \hspace{-2mm}2.56 \\
\hspace{0.6em} w/o Kw Fidelity & \hspace{-3mm}70.0 & \hspace{-3mm}12.3 &
\hspace{-3mm}71.2 & \hspace{-2mm}0.95 & \hspace{-2mm}1.19 \\
Gold & \hspace{-3mm}100 & \hspace{-3mm}100 & \hspace{-3mm}100 & \hspace{-2mm}0.78 & \hspace{-2mm}0.86 \\
\bottomrule
\end{tabular}
\caption{\small \label{tab:multiwoz-auto-eval-table} Automatic metrics
on the test set of MultiWoZ. Difference between bold and non-bold numbers is statistically significant ($p < 0.001$).}
\end{minipage}%
\hfill
\begin{minipage}{0.48\textwidth}
\centering
\begin{tabular}
{@{}lcccc@{}}
\toprule
\bf System &  \bf \hspace{-3mm}BLEU & \bf \hspace{-2mm}BRTSc & \bf \hspace{-2mm}D-2 & \bf \hspace{-2mm}ENTR\\
\midrule
KCopy & \hspace{-3mm}13.4 & \hspace{-3mm}74.3 & \bf \hspace{-2mm}3.64 & \hspace{-2mm}3.12\\
KGuide \shortcite{DBLP:conf/acl/ZhaoZE17} & \hspace{-3mm}16.7 & \hspace{-3mm}71.5 & \hspace{-2mm}2.54 & \hspace{-2mm}2.12 \\
KGround \shortcite{DBLP:journals/corr/abs-1901-08149} & \hspace{-3mm}18.3 & \hspace{-3mm}72.5 & \hspace{-2mm}2.87 & \hspace{-2mm}2.35 \\
BART \shortcite{DBLP:conf/acl/LewisLGGMLSZ20} & \bf \hspace{-3mm}19.8 & \hspace{-3mm}73.4 & \hspace{-2mm}2.97 & \hspace{-2mm}2.55 \\
RAG \shortcite{DBLP:conf/nips/LewisPPPKGKLYR020} & \bf \hspace{-3mm}19.9 & \hspace{-3mm}73.1 & \hspace{-2mm}1.03 & \hspace{-2mm}1.45\\
\ours & \bf \hspace{-3mm}19.4 & \bf \hspace{-3mm}76.8 & \bf \hspace{-2mm}3.65 & \bf \hspace{-2mm}3.44\\
\hspace{0.6em} w/o Entailment & \hspace{-3mm}18.1 & \hspace{-3mm}74.2 & \hspace{-2mm}3.17 & \bf \hspace{-2mm}3.39 \\
\hspace{0.6em} w/o Kw Fidelity & \hspace{-3mm}18.8 & \hspace{-3mm}73.3 & \hspace{-2mm}2.75 & \hspace{-2mm}2.54 \\
Gold &  \hspace{-3mm}100 & \hspace{-3mm}100 & \hspace{-2mm}2.98 & \hspace{-2mm}2.59 \\
\bottomrule
\end{tabular}
\caption{\small \label{tab:wow-auto-eval-table} Automatic metrics
on the test set of Wizard-of-Wikipedia. Difference between bold and non-bold numbers is statistically significant ($p < 0.001$).}
\end{minipage}
\vspace{0em}
\end{table*}

\subsection{Unsupervised Ranking of Candidate Final Responses}
Several previous works often over-generate and use an additional ranking step in order to select the final candidate in unsupervised text generation \cite{DBLP:conf/emnlp/QinSWBHBBC20, DBLP:conf/emnlp/ShwartzWBBC20, DBLP:conf/naacl/ParanjapeM21}. Similarly, here we want to rank the generated candidate final responses according to the diversity of the generated text as well as the conditional likelihood of generation given the dialog history. For diversity, we measure the percentage of distinct bigrams present in the response. 
For conditional likelihood, we use the pre-trained GPT2 model to obtain the log probability when the dialog history, followed by the generated response, passed as a concatenated input.
Since these two scores can have 
varied
scale, we perform Z-normalization on the individual scores and add them to obtain a single score for ranking.
The highest ranked candidate response is finally rendered to the user.

\section{Experimental Setup}

\subsection{Scenarios and Datasets}\label{sec:dataset}
We experiment with two dialog scenarios: goal-oriented and knowledge grounded. Both 
setups are knowledge intensive
but the training data in such setups often contains only a fraction of the needed knowledge. 
For the goal-oriented setting, we use the Multi-domain Wizard-of-Oz \cite{multiwoz} dataset. For knowledge grounded dialog, we use the Wizard-of-Wikipedia \cite{wow} dataset. More details are in \autoref{ap:data}.

\newpara{Multi-domain Wizard-of-Oz (MultiWOZ)} is a multi-domain dialog dataset (we use v2.0 \cite{DBLP:conf/nips/Hosseini-AslMWY20}) consisting of goal-oriented human-human conversations. 
The dataset spans 
seven domains (restaurant, train, attraction, hotel, taxi, hospital, police) and contains 10,438 dialogs with 13.68 average turns. 
Since, we do not need any training data, we only use an evaluation set (of 7K utterances).


\newpara{Wizard-of-Wikipedia (WoW)}
is a knowledge grounded dialog dataset which involves retrieving relevant knowledge from Wikipedia, reading and conditioning on it, and finally generating dialog responses \cite{wow}.
The dataset contains 201K utterances from 22K dialogues spanning 1300 diverse topics, 
from which we 
use only the test set. The associated Wikipedia knowledge base has 5.4M articles and 93M sentences.

\subsection{Baselines and Ablations}\label{sec:baselines}

\newpara{Baselines for MultiWOZ.}
For MultiWOZ, we consider several baselines following \cite{DBLP:journals/corr/abs-2010-12757} for knowledge injection. First, we use the current state-of-the-art model, SimpleTOD, for goal-oriented dialog \cite{DBLP:conf/nips/Hosseini-AslMWY20}. \citet{DBLP:journals/corr/abs-2010-12757} extends SimpleTOD by adding chitchat candidates to dialog histories during training. They also have other variants that either \textit{concatenate} output from SimpleTOD and candidate chitchats (Arranger) or \textit{rewrite} by combining both output and chitchat snippets (Rewriter). We also have a trivial baseline (KCopy) which appends the retrieved knowledge snippet $k$ from \ours{} with the initial response $x_d$.

\newpara{Baselines for WoW.}
For WoW, we use two current-best knowledge-grounded models, KGround  \cite{DBLP:journals/corr/abs-1901-08149} and BART \cite{DBLP:conf/acl/LewisLGGMLSZ20} that concatenate the associated knowledge snippets (present in WoW) and the dialog history 
as inputs to generate the response with supervision.
KGuide \cite{DBLP:conf/acl/ZhaoZE17} and RAG \cite{DBLP:conf/nips/LewisPPPKGKLYR020} have an additional knowledge selection step modeled by a latent variable before response generation similar to knowledge grounded models. We also use the KCopy baseline, as described for MultiWOZ.

\newpara{Variants of \ours.} To investigate the impact of various decoding constraints in \ours{}, we consider the following two variants of \ours---w/o Entailment and w/o Knowledge (Kw) Fidelity (\autoref{sec:decoding}). In 
\ours{},
we use SimpleTOD as the base dialog model in goal-oriented scenarios and use BART (which is a state-of-the-art model for WoW) as the base dialog model in the knowledge-grounded scenario. For all variants of \ours{}, we use gradient-based inference for decoding the final response.

\begin{table*}[t!]
\small
\resizebox{\textwidth}{!}{%
\centering
\begin{tabular}{llccc|ccc|ccc|ccc|ccc}
\toprule
& \bf \ours~vs  & \multicolumn{3}{c|}{\bf SimpleTOD} & \multicolumn{3}{c|}{\bf Rewriter} & \multicolumn{3}{c|}{\bf w/o Entailment} & \multicolumn{3}{c|}{\bf w/o Kw Fidelity} & \multicolumn{3}{c}{\bf Gold} \\ 
\cmidrule{2-17}
& \bf Criteria      & win          & loss & $\kappa$  & win          & loss & $\kappa$      &  win            & loss & $\kappa$          &  win         &  loss & $\kappa$        &  win           &  loss & $\kappa$       \\
\midrule
\parbox[t]{2mm}{\multirow{4}{*}{\rotatebox[origin=c]{90}{\bf \scriptsize{MultiWOZ}}}} & Coherent  &  \bf 93.2 & 4.4 & 0.76 & \textbf{85.6} & 10.2 & 0.75 & \textbf{98.7} & 0.8 & 0.72 & \textbf{77.8} & 17.8 & 0.78 & 26.2 & \textbf{34.4} & 0.69 \\ 
& Engaging &  \textbf{94.3} & 4.5 & 0.78 & \textbf{89.7} & 7.9 & 0.79 & \textbf{98.7} & 0.6 & 0.80 & \textbf{71.5} & 20.5 & 0.80 & 42.4 & 37.4 & 0.78\\
& Interesting &  \textbf{92.7} & 5.4 & 0.72 & \textbf{91.2} & 8.3 & 0.73 & \textbf{88.6} & 8.9 & 0.68 & \textbf{98.7} & 0.8 & 0.75 & 49.7 & 45.6 & 0.67\\
& Humanlike &  \textbf{85.4} & 10.7 & 0.68 & \textbf{87.4} & 7.3 & 0.65 & \textbf{61.9} & 30.5 & 0.71 & \textbf{81.7} & 14.0 & 0.74 & 29.7 & \textbf{37.8} & 0.66\\
\midrule
&   & \multicolumn{3}{c|}{\bf RAG} & \multicolumn{3}{c|}{\bf BART} & \multicolumn{3}{c|}{\bf w/o Entailment} & \multicolumn{3}{c|}{\bf w/o Kw Fidelity} & \multicolumn{3}{c}{\bf Gold} \\ 
\midrule
\parbox[t]{2mm}{\multirow{4}{*}{\rotatebox[origin=c]{90}{\bf WoW}}} & Coherent  &  \textbf{95.4} & 4.5 & 0.78 & \textbf{88.5} & 9.6 & 0.72 & \textbf{94.3} & 3.4 & 0.68 & \textbf{83.6} & 10.7 & 0.65 & 23.8 & \bf 25.3 & 0.73\\ 
& Engaging &  \textbf{89.3} & 7.7 & 0.72 & \textbf{87.8} & 8.3 & 0.71 & \textbf{97.7} & 0.8 & 0.70 & \textbf{71.5} & 25.4 & 0.69 & 25.4 & 26.7 & 0.73\\
& Interesting &  \textbf{96.3} & 3.5 & 0.74& \textbf{83.3} & 9.9 & 0.75 & \textbf{79.8} & 17.2 & 0.70 & \textbf{93.5} & 4.5 & 0.71 & 35.9 & \bf 37.8 &  0.76\\
& Humanlike &  \textbf{91.4} & 7.1 & 0.68 & \textbf{92.4} & 6.5 & 0.66 & \textbf{84.5} & 10.5 & 0.67 & \textbf{81.8} & 13.5 & 0.71 & 42.3 & 41.9 & 0.68\\
\bottomrule
\end{tabular}%
}
\caption{\small Pairwise comparison (\% win/loss cases, tie not reported) between responses from \ours{} and from other baselines as well as ground truth. Difference between bold and non-bold numbers is statistically significant ($p < 0.001$). $\kappa$ denotes 
Cohen's Kappa \cite{cohen1960coefficient} between a pair of annotators. 
Complete details of the human evaluation are in \autoref{ap:human}.
}
\label{tab:human-eval-table}
\vspace{-0ex}
\end{table*}

\section{Results and Discussion}
\subsection{Automatic Evaluation
}
Our primary goal is to generate responses enriched with relevant external knowledge. 
Arguably, a system which can effectively leverage additional knowledge at decoding time should generate more diverse responses.
We measure percentage of distinct bigrams as Distinct-(D-2) \cite{DBLP:conf/naacl/LiGBGD16} and geometric mean of entropy values of empirical frequency distributions of n-grams ($n = 1,2,3$) as Entropy (ENTR) \cite{jhamtani2018chess} for diversity. 
Additionally, we report overlap between generated responses and corresponding ground truth as per BLEU and BERTScore (BRTSc). For MultiWOZ, we also report the final goal accuracy (Acc) following \cite{DBLP:conf/nips/Hosseini-AslMWY20}.

\newpara{MultiWOZ.} \autoref{tab:multiwoz-auto-eval-table} shows 
\ours{}~outperforms all the baselines in terms of diversity of generated responses. More importantly, we see \ours{} promotes accuracy of reaching the final dialog state i.e.~the goal. For ablated versions of \ours{}, we find 
the 
entailment constraint has little effect on diversity while dropping 
the 
knowledge adherence constraint negatively influences accuracy and diversity. 
All variants of SimpleTOD and all versions of \ours{} show departure from the results obtained by SimpleTOD on BLEU and BERTScore since all of these versions add external knowledge 
that
were not explicitly present in the data. However, we observe that the departure is not significant and \ours{}~achieves a much closer BERTScore to SimpleTOD compared to baselines.

\newpara{WoW.}
Despite all systems for WoW use knowledge explicitly in the knowledge-grounded dialog generation task,
\autoref{tab:wow-auto-eval-table} shows \ours{}~generates the most diverse responses.
Similar to MultiWOZ, the knowledge adherence constraint still remains a significant factor for increasing diversity, one of the main goals of knowledge injection. For WoW, we instead see \ours{}~outperform even BART (previous SOTA) in terms of BERTScore when injected with external knowledge indicating the need of the external knowledge for modeling WoW dialogs. 



\begin{figure*}[t!]
    \centering
    \includegraphics[trim=95 435 145 25,clip, width=0.9\linewidth]{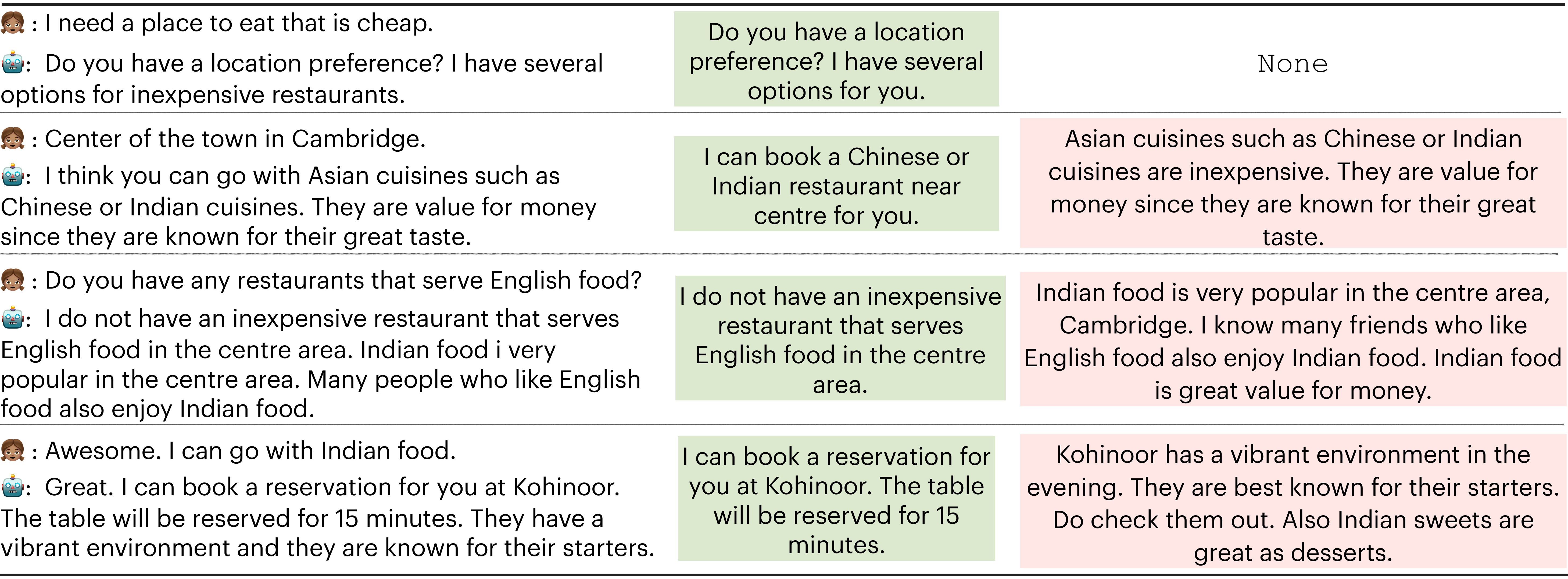}
    \vspace{-0.3em}
    \caption{\small \ours~converses with a user who is looking for some restaurant options (left column). In each turn, a \colorbox{myred}{knowledge snippet} (right column) is injected into an  \colorbox{mygreen}{initial response} (middle column).
    More examples are in \autoref{ap:example}.
    }
    \label{fig:example}
    \vspace{0em}
\end{figure*}

\subsection{Human Evaluation}
\label{sec:human-eval}

We conduct a comparative human evaluation with 300 samples to evaluate the quality of generated dialog responses
following ACUTE-Eval \cite{acuteeval}. We show a generated response from \ours{}~to an annotator with its associated dialog history to annotate if knowledge injection makes the final response more \textit{engaging}, \textit{interesting} and \textit{humanlike} compared to a baseline response. As sanity check, we also investigate if the response remain \textit{coherent} after knowledge injection. Each sample is evaluated by two annotators\footnote{\label{foot:human-eval}More details of the setup
are in \autoref{ap:human}.
}.

\newpara{MultiWOZ.}
\autoref{tab:human-eval-table} records the pairwise comparison showing \ours~consistently outperforms
baselines 
on
all
criteria.
Responses from \ours{}~are more engaging and interesting compared to SimpleTOD and Rewriter, demonstrating that gradient-based decoding is effective for knowledge injection. 
In \ours{}, entailment constraint mostly influences coherence whereas 
knowledge fidelity constraint is important for 
engagingness and interestingness.

\newpara{WoW.}
\autoref{tab:human-eval-table} shows \ours{}~outperforms 
baselines that use grounding knowledge during training in all criteria showing 
that
external knowledge can be useful even in the knowledge-grounded setting to make the conversation engaging and interesting. It also indicates the limitation of the training signal or lack of access to sufficient knowledge and room for improvement in terms of how knowledge is utilized. A large gap in win percentages in favor of \ours~for evaluating how `humanlike' is a response when compared to state-of-the-art methods suggests knowledge injection leads to more natural conversation. Here too, both decoding constraints show similar trends to MultiWOZ. 

\newpara{Qualitative Analysis.}
\autoref{fig:example} shows a conversation by \ours~with a user who seeks to find restaurant options around Cambridge.
We observe that in most of the turns the injected knowledge appeared as an additional justification over the initial responses making the dialog engaging and effective to reach the user's goal (also noted
by human judges in \autoref{sec:user-study}).
For example, in turn 3, we observe that adding the extra information about Indian cuisine helped user to reach a conclusion when their original choice of English cuisine was absent.

\newpara{Effect of Response Length.}
Qualitatively, as seen in \autoref{fig:example}, responses generated by \ours~are longer than those from the initial response due to the post-hoc knowledge injection. In the human evaluation sample, we found that 37\% of responses from \ours~are similar or smaller in length compared to responses from the best baseline. We investigate if response length acted as a confounding factor during human evaluation. Among all the cases where \ours~was \emph{lost} over a baseline, 45\% ($\pm$~2\% when bootstrapped with 1000 subsets of size 50) of responses from \ours~were longer than those from the comparing baseline. Among \emph{win} cases for \ours, we observe 49\% ($\pm$~3\% when bootstrapped with 1000 subsets of size 50) \ours~responses were longer than those from the comparing method. This indicates that human users did not only choose longer responses as better.

\subsection{User Study for Effectiveness of Knowledge Injection}
\label{sec:user-study}

Relevant knowledge injection has the benefit of adding more justification to terse dialog outputs and hence influencing the task outcome positively. 
Mirroring observations from \cite{DBLP:conf/nips/GhandehariounSJ19}, a real-time full conversation evaluation is needed to investigate if \ours~could achieve the conversational goal any better than 
baselines.

We recruited 60 
users for this study\footnote{More details of the participants and the study setup are in \autoref{ap:human}.}. One half of the users interacted with \ours{}, while the other half interacted with the best baseline model that does not augment dialog responses with external knowledge. We construct a \textit{speculative goal} for each user to accomplish via the conversation.
We allow users to end the conversation any time they would like and ask them whether 
the system helped them to reach their conversation goal along with additional comments to
justify their annotation. Users 
who interacted with a knowledge-augmented system
also asked if the system provided any knowledge that user has not explicitly asked for but indeed the extra information helped them to reach the conversational goal \cite{DBLP:journals/corr/abs-2104-06828}. Finally, we also ask if they would
like to engage with 
the system they interacted with in future.

For goal-oriented dialog, we construct speculative goals (e.g.~looking for entertainment options) manually from the ground truth for 300 dialog samples. Since we are not using the underlying databases, we made sure speculative goals do not require specific information (e.g.~booking availability, flight information, etc.). For knowledge-grounded dialog, we provide the intended topic of discussion (e.g.~science fiction) present in the data; 
the speculative goal here is to know more about, or to have an engaging conversation about the topic.

\begin{table}[t!]
\small
\centering
\resizebox{0.95\linewidth}{!}{%
\begin{tabular}{@{}lccccc@{}}
\toprule
\bf MultiWOZ & \bf \# turns $\downarrow$ & \bf Goal & \bf Know & \bf Would use\\ \midrule
\hspace{0.7em} Rewriter & 8 $\pm$ 2 & 69\% & 35\%         & 56\%      \\ 
\hspace{0.7em} \ours{}  & \bf 4 $\pm$ 3 & \bf 86\% & \bf 84\%  & \bf 76\%      \\
\midrule
\bf WoW & \bf \# turns $\uparrow$ & \bf Goal & \bf Know & \bf Would use\\ \midrule
\hspace{0.7em} BART & 10 $\pm$ 2 & 56\% & 70\%         & 48\%      \\ 
\hspace{0.7em} \ours{}  & \bf 16 $\pm$ 3 & \bf 76\% & \bf 89\%         & \bf 71\%      \\
\bottomrule
\end{tabular}%
}
\vspace{-0.3em}
\caption{\small Real-time user study with average \# of turns for successful goal completion, \% of time the goal was achieved, \% of success cases users were helped by an \textit{additional} knowledge (Know) that was not explicitly asked to reach their goal, and if users would like to use the system in future.}
\vspace{-0.3em}
\end{table}

\newpara{Results.}
First of all, we find that \ours{}~is unanimously preferred by users compared to the baseline during the user study.
More importantly, we see that when the user successfully accomplished their goal, 84\% of those times they found the additional knowledge helpful in the goal-oriented setting (MultiWOZ) as compared to a baseline (Rewriter) that did not use any external knowledge. Most importantly, \ours{}~takes significantly 
fewer
turns for users to accomplish the goal as compared to Rewriter implicitly indicating injected knowledge (we observe high correlation, 0.67) contributes toward more efficient conversations. 

For the knowledge-grounded setting (WoW), both BART and \ours{}~have access to external knowledge sources. However, 89\% (compared to 70\%) of success scenarios were directly influenced by the additional post-hoc knowledge. For knowledge-grounded dialog, a longer conversation is indicative of engagingness on a particular topic \cite{DBLP:conf/interspeech/GopalakrishnanH19}, hence users preferred to converse with \ours{}~for more
turns as compared to a BART baseline. We quote a comment from a user who found a conversation about the Korean culture with \ours{}~was particularly engaging---``\textit{Before this conversation, I had less knowledge about Korean movies and art-forms. This gave me a new perspective and a handful of popular opinions to look at it.}''.

\subsection{Discussion}

\begin{table}[t!]
    \centering
    \resizebox{\linewidth}{!}{%
    \begin{tabular}{@{}lcccccc@{}}
    \toprule
     &  \multicolumn{2}{c}{\bf Relevant} & \multicolumn{2}{c}{\bf Factual} & \multicolumn{2}{c}{\bf BRTSc for WoW}\\
    \cmidrule(l{2pt}r{2pt}){2-3} \cmidrule(l{2pt}r{2pt}){4-5} \cmidrule(l{2pt}r{2pt}){6-7}
    \bf Source & \bf Random & \bf DPP & \bf Random & \bf DPP & \bf Random & \bf DPP\\
    \midrule
    Parametric           &   82\%  &  \bf 89\%  &   65\%   & \bf 83\%  &  74.2  &   \bf 81.3 \\
    Non-parametric      &  81\%  & \bf 83\%  &  97\%  & \bf 98\% & 65.2 & \bf 76.8\\
    \bottomrule
    \end{tabular}%
    }
    \vspace{-0.3em}
    \caption{\small Evaluation for the quality of the knowledge snippets for random and DPP-based selection.}
\label{tab:know-select}
\vspace{0em}
\end{table}

\begin{table}[t!]
\small
\centering
\resizebox{0.90\linewidth}{!}{%
\begin{tabular}{@{}lrr@{}}
\toprule
\bf System     & \bf MultiWOZ            & \bf WoW      \\ \midrule
Supervised & 17.6 $\pm$ 5.2 ms & 23.6 $\pm$ 4.6 ms\\
PPCM \shortcite{DBLP:conf/emnlp/MadottoILDF20}  & 30.9 $\pm$ 7.5 ms & 32.6 $\pm$ 4.2 ms \\
\ours{} & 34.2 $\pm$ 8.4 ms & 35.7 $\pm$ 5.7 ms \\
\ours{}, only decoding & 31.6 $\pm$ 2.7 ms & 32.3 $\pm$ 3.4 ms\\
\bottomrule
\end{tabular}%
}
\vspace{-0.5em}
\caption{\small Mean and std.~error of clock-time taken per token}
\label{tab:timings}
\vspace{0em}
\end{table}

\newpara{Performance of Knowledge Selection.}~
The knowledge selection step in \ours{}~acts an information bottleneck where the quality of the generated response directly depends on the quality of the selected knowledge\footnote{A statistical analysis on number of knowledge snippets retrieved/generated and selected is provided in \autoref{ap:code}.}. We perform a human evaluation on 200 snippets to measure the relevance and the factual correctness in two scenarios: when we randomly select a retrieved snippet or select via DPP.
In \Cref{tab:know-select}, we see that 
the parametric knowledge source (\texttt{gpt2-large}) generates more relevant knowledge snippets than a non-parametric one. We attribute this to 1) a large and diverse dataset (webtext) used during pretraining of \texttt{gpt2} as compared to yelp reviews (restricted domains) we used for retrieval,
and 2) the limited recall of relevant knowledge when using word-overlap based retrieval. However, large language models are still prone to generate non-factual knowledge.
We observe that DPP-based selection in \ours{}~is able to sub-select more factual knowledge which then positively influences the final response quality. For WoW, we also compare the selected snippets with the gold knowledge available in the dataset that in turn show high fidelity in terms of BERTScore.

\newpara{Time Complexity.~}
\citet{DBLP:conf/emnlp/MadottoILDF20}~shows that iterative gradient-based decoding could be slower than generating response using single forward pass from an existing model. When we benchmark \ours{}~in an Nvidia 2080Ti GPU, in \Cref{tab:timings},
we see that knowledge generation (or retrieval) could be a computational bottleneck for \ours{}. However the greedy selection and the constrained decoding step do not add significant computational load. Furthermore, \ours{}'s performance is comparable with PPCM \cite{DBLP:conf/emnlp/MadottoILDF20}---a more efficient version of gradient-based decoding.
The efficiency of the knowledge retrieval step can be improved with better indexing \cite{DBLP:journals/tbd/JohnsonDJ21} which we leave as a future work.

\section{Related Work}

Knowledge grounded dialog datasets such as Wizard-of-Wikipedia \cite{DBLP:conf/iclr/DinanRSFAW19} and Topical chat \cite{DBLP:conf/interspeech/GopalakrishnanH19} typically 
consist of
dialog responses paired with relevant knowledge available as collected annotations.
Hence, models trained on such datasets are 
restricted to the knowledge sources they were exposed to at training time.
Past work 
\cite{DBLP:journals/corr/abs-2010-12757, DBLP:conf/emnlp/MajumderJBM20,su2020diversifying, DBLP:journals/corr/abs-2107-07566, DBLP:journals/corr/abs-2111-05204, DBLP:conf/aaai/GhazvininejadBC18,DBLP:journals/corr/abs-2004-14614, rag2020, realm2020}
has
looked into injecting extra knowledge sources 
at
training time in a bid to add knowledge not available originally as paired to dialog responses.
However, such approaches require re-training the model if some new knowledge source were to be used. Moreover, while previous work focuses on just improving specificity of dialog response using external knowledge, we also study the effect of additional knowledge in achieving conversational goals.

Improving the diversity of dialog responses by using diversity-promoting sampling has been explored in past work \cite{DBLP:conf/acl/LewisDF18,DBLP:conf/iclr/HoltzmanBDFC20}.
We use a gradient-based decoding method, building on past work in this direction \cite{DBLP:conf/iclr/DathathriMLHFMY20,DBLP:conf/emnlp/QinSWBHBBC20,DBLP:conf/emnlp/MadottoILDF20, DBLP:conf/acl/MajumderBMJ20}. However, we propose new objectives to inject post-hoc knowledge obtained based on already generated dialog---an unsupervised knowledge injection method that has not been explored so far.  


\section{Conclusion}
We propose a framework for unsupervised knowledge injection into dialog responses. We show that knowledge can be obtained post-hoc from \textit{any} knowledge sources that can improve users' ability to reach their conversational goal more effectively. In future, our idea can be generalized to setups 
where external knowledge can justify model's predictions
such as conversational recommendation.

\section*{Acknowledgements} We thank anonymous reviewers for providing valuable feedback. BPM is partly supported by a Qualcomm Innovation Fellowship, a Friends of the International Center Fellowship--UC San Diego, NSF Award \#1750063, and MeetElise. 
\bibliography{anthology,custom}
\bibliographystyle{acl_natbib}


\appendix
\section{Datasets}
\label{ap:data}

\paragraph{MultiWOZ.} 
To compare with previous works, we 
use
MultiWoz 2.0 
following 
\cite{DBLP:conf/nips/Hosseini-AslMWY20}. Note that we do not need any training data for our models since we perform post-hoc knowledge injection.

\paragraph{WoW}
For Wizard-of-Wikipedia, all baselines and the original dialog model for \ours{}~use available paired knowledge present in the training data (not a part of our pipeline). However, \ours{}~additionally uses the external knowledge snippets selected via DPP.  

\section{Implementation Details}
\label{ap:code}

We open-source our code at: \url{https://github.com/majumderb/poki}. We use the publicly available implementation\footnote{\url{https://github.com/guilgautier/DPPy}} for DPP \cite{GPBV19}. 

We obtain the MultiWOZ 2.0 from the official release \footnote{\url{https://github.com/budzianowski/multiwoz}}. Similarly, we obtain the Wizard-of-Wikipedia from ParlAI repository \footnote{\url{https://parl.ai/projects/wizard_of_wikipedia/}}.
We adapted codes from original PPLM \cite{DBLP:conf/iclr/DathathriMLHFMY20} repository\footnote{\url{https://github.com/uber-research/PPLM}} and modified them for our own objective function. 
We obtained the Yelp review dataset from the official website\footnote{\url{https://www.yelp.com/dataset}}. Yelp dataset contains 8,635,403 reviews. 
For diversity calculation (in automatic evaluation), we use NLTK\footnote{\url{https://www.nltk.org/\_modules/nltk/util.html}} to extract n-grams.

\paragraph{Network architecture}
For MultiWOZ, we use the SimpleTOD\footnote{\url{https://github.com/salesforce/simpletod}} as the base model.
Whereas for WoW, we use BART\footnote{\url{https://huggingface.co/transformers/model_doc/bart.html}} as the base model.
For the parametric knowledge source, we use \texttt{gpt2-large}\footnote{\url{https://huggingface.co/transformers/model_doc/gpt2.html}}.

\paragraph{Hyperparameters}
\ours{}~does not require any training since we perform gradient-based decoding at the inference time. For hyperparameters involved in the decoding stage, we maximize the Z-normalized sum of dialog model perplexity and linguistic diversity (\% of distinct bigrams) of the generated response in a greedy fashion to select the best values. For our best method, in objective function $\mathcal{L}$, we use $\alpha$ as 1 and $\lambda$ as 1. We keep generation length to be 100 to encourage longer generations. We train the entailment classifier using code from PPLM repository\footnote{\url{https://github.com/uber-research/PPLM/blob/master/run_pplm_discrim_train.py}}. The weight $\gamma$ for mixing forward and backward passes was set to 0.45. We run 5 backward-forward passes to obtain a candidate final response. 

\paragraph{Filtering knowledge candidates from PTLMs}
Our initial experiments suggests that that knowledge generated from PTLMs can be inappropriate (contains bias or toxic content) and misleading/nonfactual. 
\citet{DBLP:journals/corr/abs-2010-12757} collected annotations of dialog responses with labels \texttt{positive} (useful, social), \texttt{negative} (inappropriate and misleading). We learn a binary classifier to classify a knowledge snippet as positive or negative and use it as a filtering criteria. 

\paragraph{Key-phrase extraction}
Given a sentence from the context, we first extract n-gram (n $\in$ 1,2,3,4) key-phrases using YAKE (Yet-Another-Keyword-Extractor) \cite{campos2020yake} and retain only those that contain at least a noun. 

\paragraph{Prompts}
We curated prompts inspired by various knowledge-seeking situations (such as for: more information, opinion, review) \cite{DBLP:conf/emnlp/ShwartzWBBC20} and are listed in \autoref{tab:prompts}.

\begin{table}[h!]
\small
    \centering
    \begin{tabular}{l}
    \toprule
    \texttt{[KP]} is famous for \rule{0.5cm}{0.15mm}  \\
    The popular opinion about \texttt{[KP]} is \rule{0.5cm}{0.15mm}\\
    Here is what I know about \texttt{[KP]}: \rule{0.5cm}{0.15mm}\\
    My friend says that \texttt{[KP]} is: \rule{0.5cm}{0.15mm}\\
    Here is some information about \texttt{[KP]}: \rule{0.5cm}{0.15mm}\\
    Here are some reviews about \texttt{[KP]}: \rule{0.5cm}{0.15mm}\\
    I think \texttt{[KP]} is: \rule{0.5cm}{0.15mm}\\
    I read on the internet about \texttt{[KP]} and found that \rule{0.5cm}{0.15mm}\\
    Today I learned about \texttt{[KP]} that \rule{0.5cm}{0.15mm}\\
    \bottomrule
    \end{tabular}
    \caption{Manually curated prompts to query the PTLM}
    \label{tab:prompts}
\end{table}

\paragraph{Statistics on generated and selected knowledge snippets} For both datasets, we retrieve 100 most relevant knowledge snippets from non-parametric source (here, yelp reviews), and generate 5 candidate knowledge snippets (using nucleus sampling \cite{DBLP:conf/iclr/HoltzmanBDFC20}, $p=0.95$) for each key-phrase extracted from an input instance (dialog history + initial response). After knowledge selection by DPP, on an average (over validation set), 5 snippets were selected for MultiWoz and 8 snippets were selected for WoW.   

\section{Human Evaluation and User Study Setup}
\label{ap:human}

\begin{figure*}[t]
    \centering
    \includegraphics[width=\linewidth]{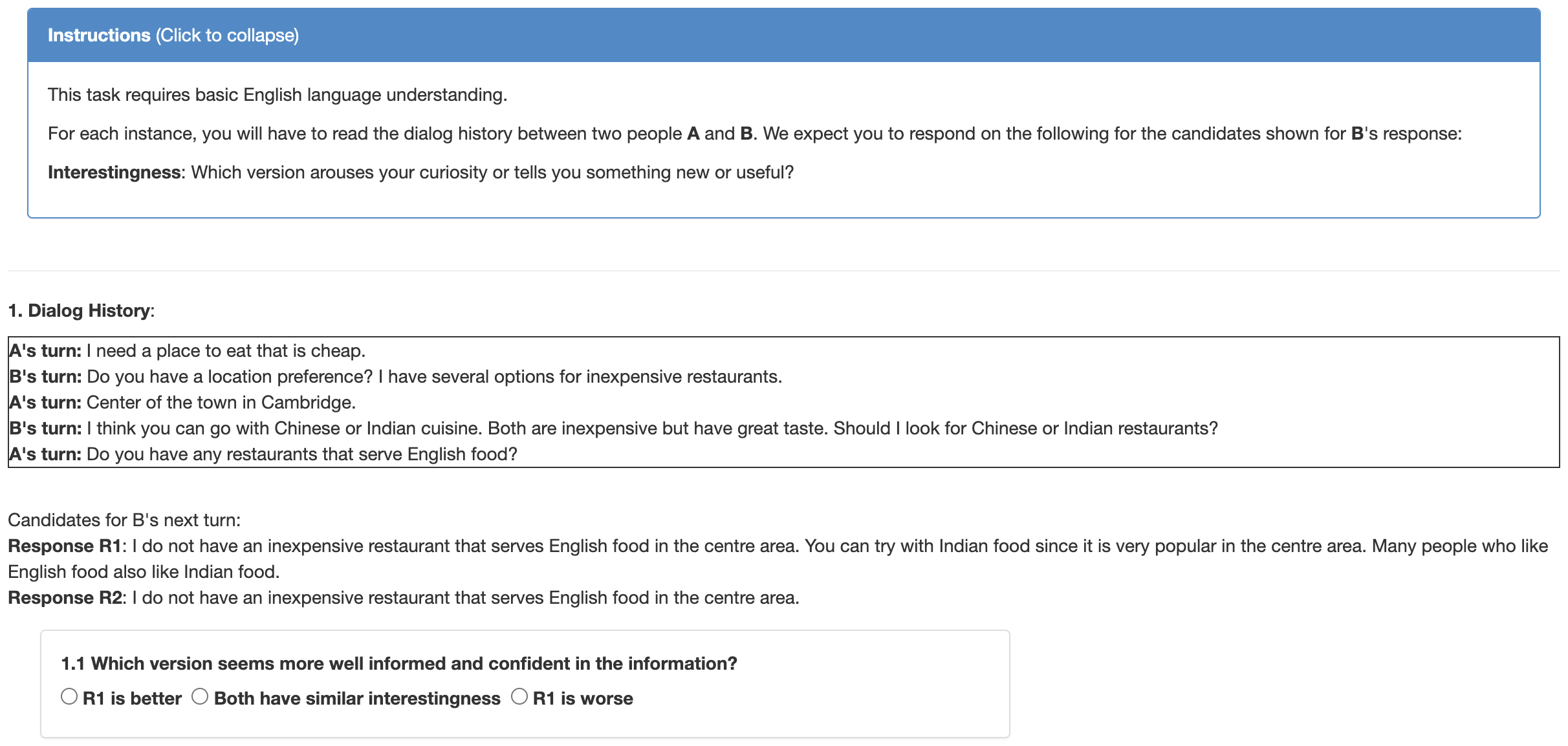}
    \caption{Human evaluation setup for pairwise comparison between \ours{}~and another baseline}
    \label{fig:human_eval1}
\end{figure*}

\paragraph{Human Evaluation}
We hired two Anglophone (Lifetime HIT acceptance \% $>$ 85) annotators for every test sample. \Cref{fig:human_eval1} shows a sample question for the pairwise comparison between response generated by \ours{} and a baseline for informativeness. The exact formulations for all criteria are provided as below:
\begin{itemize}[noitemsep]
    \item \textbf{Coherent}: Which version is more consistent with the dialog history?  
    \item \textbf{Engaging}: \textit{Which version is more likely to hold your attention and make you want to hear more?}
    \item \textbf{Interesting}: \textit{Which version arouses your curiosity
    or tells you something new or useful?}
    \item \textbf{Humanlike}: \textit{Which version is more natural and personable?}
\end{itemize}

All differences in values from human evaluations are significant with $p < 0.05$ from bootstrap tests on 1000 subsets of size 50. A snapshot of our human evaluation interface is shown in \autoref{fig:human_eval1}. The order of two candidate responses (R1 and R2) is made random for each question. 

\paragraph{User Study}
For user study, we similarly recruited 60 Anglophone users who have at least high-school level of education and are comfortable with handling internet-based technologies. Each session (depending on the systems they interacted) lasted on an average 30 minutes (for MultiWOZ) and 60 minutes (for WoW) including on-boarding, performing actual task and answering post-task questions.

\begin{figure*}[t]
    \centering
    \includegraphics[trim=450 400 450 70,clip,width=0.9\linewidth]{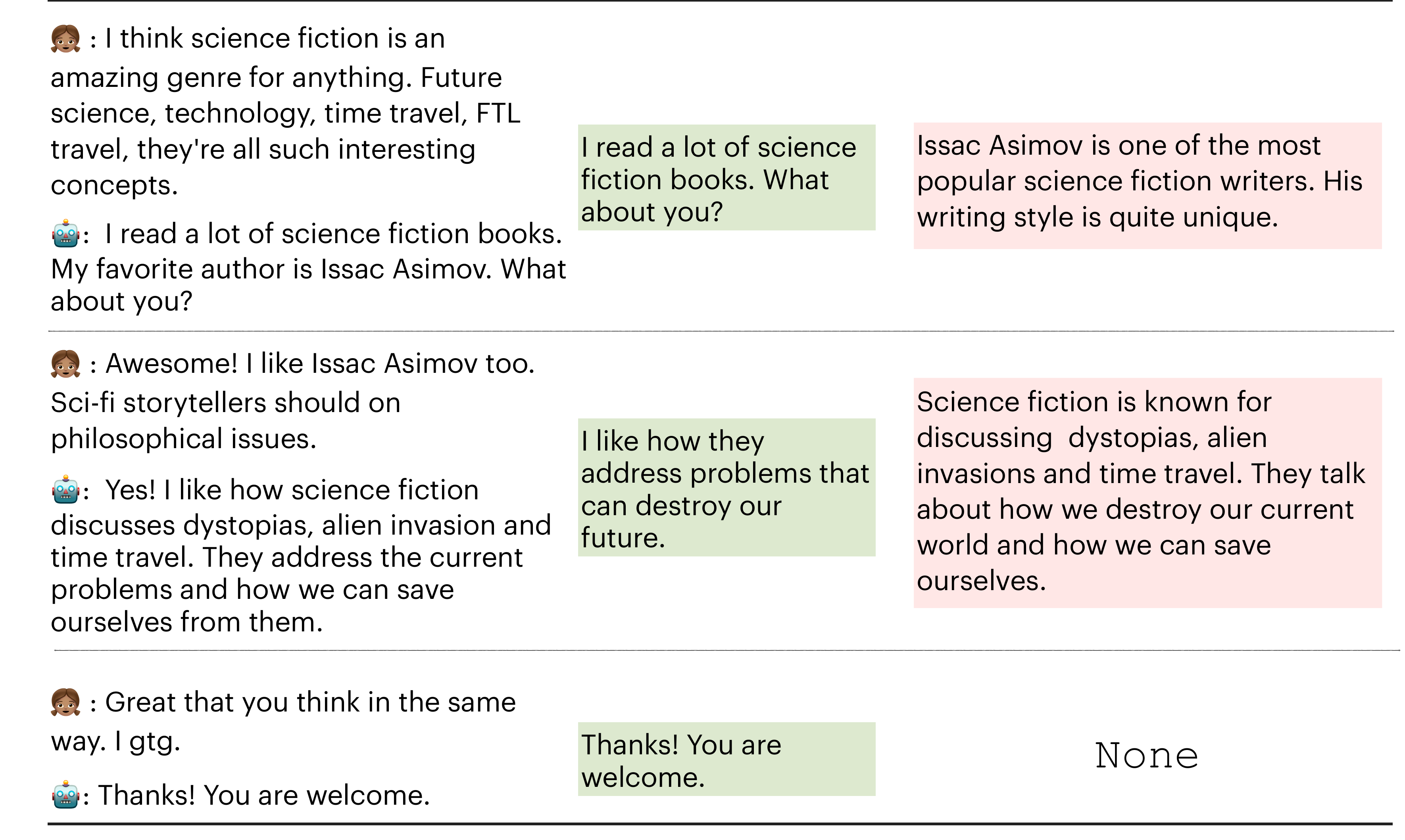}
    \caption{ \ours~converses with a user who is discussing about science fiction, in a knowledge-grounded dialog scenario (left column). In each turn, a \colorbox{mygreen}{initial response} (middle column) is augmented with a \colorbox{myred}{knowledge snippet} (right column) using constrained gradient-based decoding. Human judges unanimously noted this conversation as more engaging as compared to the initial responses.}
    \label{fig:kg-ex}
\end{figure*}

\begin{figure*}[t]
    \centering
    \includegraphics[trim=450 390 450 60,clip,width=0.9\linewidth]{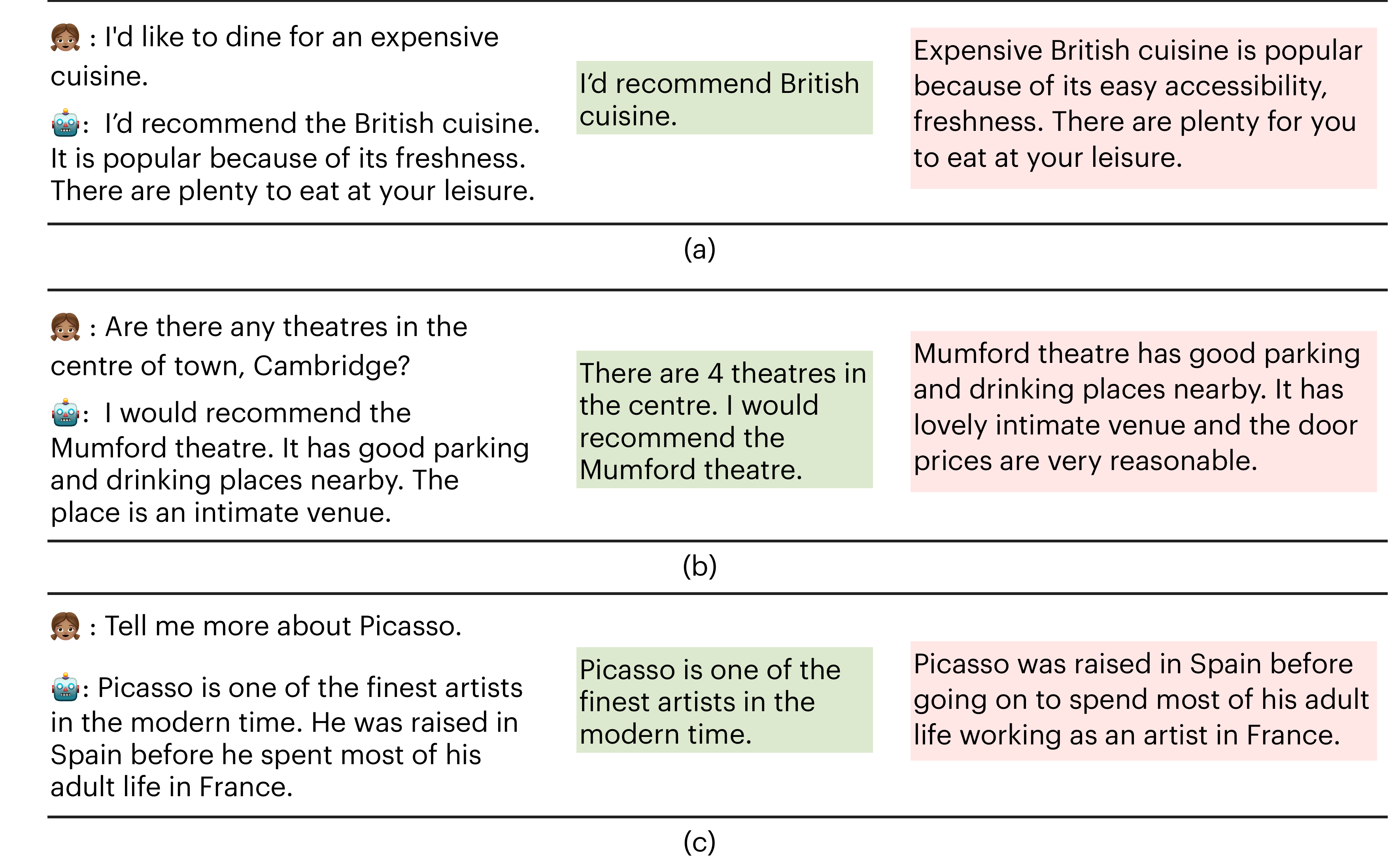}
    \caption{Utterance level examples (left column) in (a) and (b) goal oriented scenario; and (c) knowledge-grounded scenario. \ours~updates the  \colorbox{mygreen}{initial response} (middle column) with a \colorbox{myred}{knowledge snippet} (right column) using constrained gradient-based decoding.
    }
    \label{fig:ut-ex}
\end{figure*}


\section{Qualitative Examples}
\label{ap:example}

\autoref{fig:kg-ex} shows a complete dialog in the knowledge-grounded scenario where the user discusses about `science-fiction'. \autoref{fig:ut-ex} shows more utterance level examples for both goal-oriented and knowledge-grounded scenarios.

\section*{Ethical considerations}

We do not foresee any immediate ethical concerns for our method as we use several constraints (less divergence from the extracted knowledge, consistency with the dialog context) that allow the generation to be restricted to the context. In general, we expect our dialog system to be more engaging and accessible to the user. 
Since we use PTLMs as knowledge source, we inherit the general
risk of generating biased or toxic language, which should be carefully filtered. In our work, we perform explicit filtering steps to make sure that the knowledge is \textit{appropriate}. Furthermore, our selection step promotes more factually correct knowledge to be selected. However, the generations may incorporate biases that are already present in the dialog datasets due to crowd-sourced data collection.
Finally, our generations are limited only to the English language.
Hence we suggest that a system like ours should likely not be used as a `black box,' but would best be used in a setting where its outputs can be `audited'. \textbf{Carbon footprint:} Our system uses post-hoc knowledge injection which refrains from retraining newer dialog models to accommodate dynamically evolving external knowledge. This promotes green NLP applications \cite{DBLP:journals/cacm/SchwartzDSE20,DBLP:conf/acl/StrubellGM19} reducing carbon footprints that stem from training (or even finetuning) large language models.



\end{document}